\documentclass{article}
\usepackage{graphicx}
\usepackage{array}
\usepackage{tabularx}
\usepackage{booktabs}
\usepackage{enumitem}
\usepackage{amsmath}
\usepackage{amssymb}
\usepackage{float} 
\usepackage[utf8]{inputenc}
\usepackage[T1]{fontenc}
\usepackage{url}
\usepackage{amsfonts}
\usepackage{nicefrac}
\usepackage{microtype}
\usepackage{xcolor}
\newcommand{\ub}{\_\allowbreak}
\newcommand{\arr}{$\rightarrow$}

\newcolumntype{L}{>{\raggedright\arraybackslash}X}
\newcolumntype{C}{>{\centering\arraybackslash}X}
\newcolumntype{A}{>{\raggedright\arraybackslash\hsize=1.25\hsize}X}
\newcolumntype{B}{>{\raggedright\arraybackslash\hsize=0.75\hsize}X}

\usepackage[preprint]{neurips_2026}
\usepackage{caption}
\usepackage{hyperref}
\makeatletter
\renewcommand{\@noticestring}{Accepted at the Third Workshop on GenAI for Health (GenAI4Health), NeurIPS 2026. Non-archival.}
\makeatother

\title{Objectify, Don't Delete: Complaint-Derived Evaluation of Ambient Clinical Notes}
\workshoptitle{The Third Workshop on GenAI for Health: Agentic Systems, Clinical Trust, and Future Potential}

\author{%
  \textbf{Nikita Gupta,} \textbf{Narayanan Unny Edakunni,} \textbf{Vikram Okaly,}\\
  \textbf{Suranjana Datta,} \textbf{Navya Penubadi,}\\
  \textbf{Karl Eric Swanson,} \textbf{Siddharth Sharma}\\[0.4em]
  Suki AI, California, USA%
}

\begin{document}

\maketitle

\begin{abstract}
Evaluating the quality of AI-generated clinical notes presents a fundamental challenge: standard metrics like ROUGE measure lexical similarity to reference notes, while expert checklists encode what committees believe matters. Neither captures what clinicians actually complain about. Recent work has shown that evaluation criteria can be mined from user feedback and filtered to criteria that LLM judges can reliably enforce. We argue that this is where much of the value is lost. The criteria that fail enforceability are disproportionately the ones clinicians raise most often, because high-frequency complaints are stated as several questions glued together. No yes/no judgment can grade such composites as posed. We introduce an objectification step that rewrites such criteria into sets of scoped, determinate decisions. We demonstrate the approach through an evaluation pipeline that mines criteria from complaints, objectifies them, and deploys LLM judges against both planted defects and production notes. Across eleven defects planted into ACI-bench reference notes, objectified judges localize violations with median recall of 0.95, while reference metrics change by at most 0.037. Given the complaint as one broad question, the same judges still find every planted defect, but they also flag 70\% of the unperturbed reference notes. Splitting that complaint into one section-pair question per call raises Youden's $J$ from 0.30 to 0.72; with prompt effort matched, the per-check gain is 0.19--0.26 on three models, and the single broad question remains the better note-level screen. We release ACI-Defect, a perturbation benchmark; clinician-reviewed instance-level labels showing quality gaps in the ACI-bench gold standard (32 of 40 notes contain hallucinations; 40 of 40 omit discussed facts); and grounding-judge verdicts. On 1,446 production notes, mean instance-rate burden falls at every clinician-rating step ($\rho{=}{-}0.15$). The association is small, and a named complaint does not track its mapped judge. Defect counts have to be kept per span rather than collapsed into one pass/fail bit per note.

\end{abstract}

\section{Introduction}

AI-powered ambient clinical documentation systems transcribe a patient-clinician encounter and draft a structured clinical note. They are now deployed at scale. As with any AI-generated content, it is essential that the quality and clinical safety of the generated notes be rigorously monitored. Manual review of these notes does not scale, has high inter-rater variance, and cannot be run on every production note within a short span of time. Hence, focus has shifted to automated methods of measuring quality, one of which is by comparing a note to a {\em gold} note \citep{yim2023acibench}, termed \textit{reference-based measurement}. This form of measurement fails due to the subjectivity in constructing a gold note. There is not just one perfect gold note. Two clinicians documenting the same encounter produce materially different, equally acceptable write-ups. Lexical and embedding similarity to a single reference is the protocol followed generally in practice (as demonstrated in ACI-bench \citep{yim2023acibench} which reports ROUGE \citep{lin2004rouge}, BERTScore \citep{zhang2020bertscore}, BLEURT \citep{sellam2020bleurt}, and MedCon \citep{vanveen2024adapted}). However, while it penalizes legitimate stylistic variations, it remains insensitive to clinically harmful errors that barely change $n$-gram overlap, such as a fabricated normal exam finding or an Assessment sentence leaking into the History of Present Illness.

Alternatively, instruments such as PDQI-9 \citep{stetson2012pdqi}, PDSQI-9 \citep{croxford2025pdsqi}, QNOTE \citep{burke2014qnote}, and similar checklists measure the quality of notes taking into account what is clinically relevant, and measure the impact on clinical safety. Measuring these metrics on clinical notes often requires manual examination to carefully assess the quality along different dimensions. These measures tend to assess the quality at a macroscopic level, judging the note as a whole rather than its constituents. This tends to introduce subjectivity into the measurement. There are systems \citep{croxford2025llmjudge} that try to automate the measurement of these metrics using LLM-as-a-judge, but the resultant judge also tends to be subjective and non-deterministic to a large extent. The other shortcoming of these {\em expert} metrics is that they are agreed upon by a group of experts focusing exclusively on the quality of the notes. In practice, users care less about macroscopic scores than about specific deficiencies that force them to edit the note, which adds to burnout. Table~\ref{tab:instruments} (Appendix~\ref{app:instruments}) makes the gap concrete. Recent work \citep{zhou2025feedback} mines metrics from user feedback but discards many of the most common pain points because they are composite (Table~\ref{tab:c1-themes}).

In this paper, we address these challenges with a methodology that covers the lifecycle of automated measurement of note quality: (1) mine the predominant user complaints; (2) break them down into metrics for the note and its constituents (objectification); (3) judge those metrics with an LLM; and (4) use an LLM--human adjudication loop to correct the labels and to separate a judge error from a human error.

\begin{figure}[t]
  \centering
  \includegraphics[width=\linewidth]{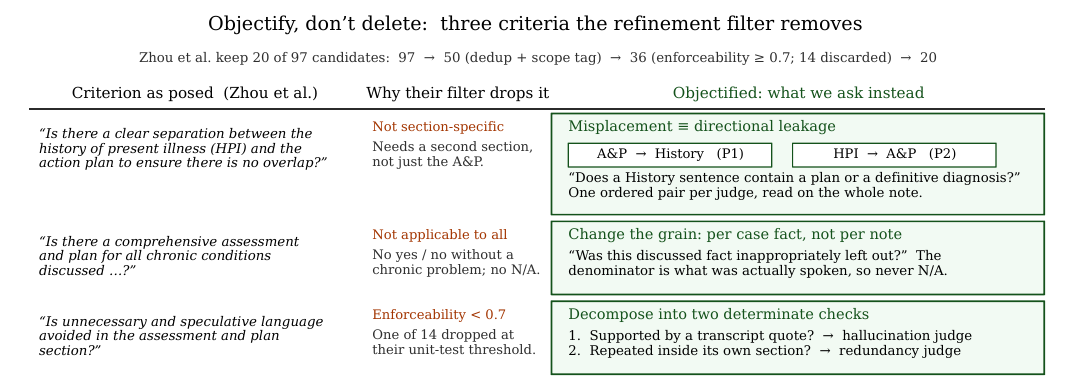}
  \caption{Objectification keeps the criterion and changes the question. Three items \citet{zhou2025feedback} drop (quoted from their appendix) become
  directional leakage, per-fact omission, and a pair of determinate checks. Their filter is correct under a one-section, no-N/A scope; we change that
  scope rather than delete the item.
}
  \label{fig:objectification}
\end{figure}

\section{Related Work and Contributions}
In this section, we analyze prior work and list our contributions over and above the state of the art.

\textbf{Feedback-derived evaluation criteria:} \citet{zhou2025feedback} turn 22K encounters of clinician free-text feedback and ratings from a deployed ambient scribe into yes/no checklists, refined by de-duplication, applicability tagging, LLM-enforceability unit tests, and beam-search subset selection. They score notes with an LLM judge and train a logistic regression on binarized ratings using checklist questions as features. We treat this pipeline as the starting point and address a shortcoming of their enforceability-filtering step. In their work, they discard user feedback that cannot be turned into a yes/no question for the LLM to judge. This forgoes the chance to create judges for potentially impactful quality gaps. We instead break such feedback down into criteria that can be judged objectively by an LLM judge. We call this the \emph{objectification} step. Examples are shown in Figure~\ref{fig:objectification}. Outside the clinical domain, SPUR \citep{lin2024spur} induces satisfaction rubrics from user data and uses rubric scores as classifier features. Feedback-to-Rubrics \citep{yoshida2026feedback} and PARL \citep{qiu2026parl} induce and iteratively refine rubrics from comments and user history. Both SPUR and \citet{zhou2025feedback} derive features from the user’s own words or from checklist questions about those words, and therefore cannot score an item for which no feedback exists.

\textbf{LLM-as-a-judge for clinical text:} PDSQI-9 \citep{croxford2025pdsqi} and its LLM-as-a-judge instantiation \citep{croxford2025llmjudge} report high agreement with expert raters on summarization quality. Checklist-based judging (CheckEval, RocketEval, HypoEval \citep{lee2025checkeval,wei2025rocketeval,li2026hypoeval}) improves reliability over single holistic scores. However, these judges mimic human evaluators on macroscopic characteristics of quality rather than on the localized constituents of the note. We instead score those constituents and aggregate them into a note-level view. Evaluating the judge itself is a separate problem: \citet{zheng2023mtbench} show that LLM judges disagree with humans in systematic ways, and \citet{shankar2024validators} show that an LLM-written rubric can look reliable while failing the people it is meant to stand in for. We therefore test localization on planted spans and treat the production rating as a weak external check, not as validation of the judge.

\textbf{Disagreement and label noise:} Concord \citep{bliss2026concord} uses inter-judge agreement for disagreement-driven triage, multi-LLM disagreement has been used to detect human annotation errors in clinical free-text \citep{wittlinger2026multillm}, and label-error detection with LLM ensembles is an active research topic \citep{nahum2025llms}. Our observation is that an interactive LLM--human adjudication loop serves two ends: (1) it corrects the labeled corpus where the judge is right (Table~\ref{tab:c2}), and (2) it localizes the judge's error to a cause specific enough to improve the prompt where the human is right. We demonstrate the first directly and use the second to specify judge revisions (we do not report a re-measured judge after revision).

\textbf{Generation benchmarks:} ACI-bench \citep{yim2023acibench} is the largest public corpus of full encounter dialogues paired with expert notes, and the de facto generation testbed (MEDIQA-Chat / MEDIQA-SUM).  PriMock57 \citep{papadopoulos2022primock57} and MTS-Dialog \citep{benabacha2023mtsdialog} are related but either much smaller or snippet-level. In this paper, we evaluate ACI-bench using human experts and AI judges and release the labels. We also release a dataset with planted quality defects for evaluating note-quality judges.

\begin{figure}[H]
  \centering
  \includegraphics[width=\linewidth]{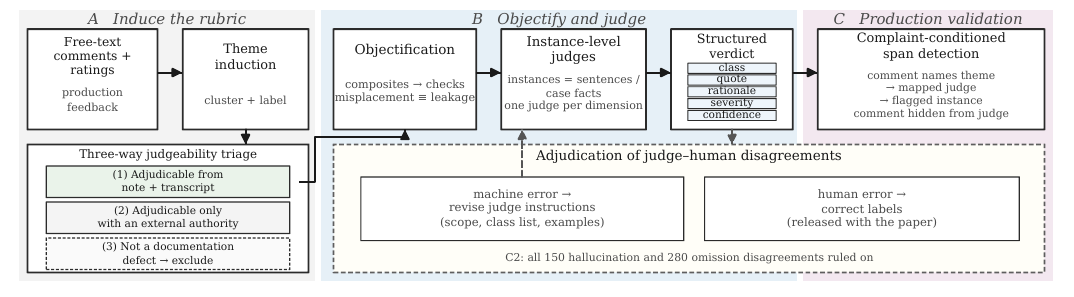}
  \caption{Evaluation loop. Comments induce the rubric (C1). Judges score sentences or case facts, not the note as a whole. Production ratings are compared to those instance rates; comments and ratings are hidden from judges. Deployed prompts, orchestration, and production notes are not released.}
\label{fig:method}
\end{figure}

\subsection{Contributions}
We summarize our contributions below and refer to them as C1–C4 in the rest of the paper.\footnote{Code, criterion strings, planted notes, and ACI-Defect labels: \url{https://github.com/nikita0511/aci-defect}.}

\textbf{C1:} We characterize the \textit{enforceability} gap in feedback-derived rubrics and introduce \textit{objectification}.

\textbf{C2:} We show that ACI-bench gold is not a clean negative set under a complaint-derived rubric, and we release independent instance-level clinical labels on dimensions the corpus does not score, together with grounding-judge precision and recall against those labels.

\textbf{C3:} We release \textit{ACI-Defect}, a perturbation overlay on the ACI-bench dataset that plants specific, clinically impactful defects rather than generic corruptions. We compare objectified judges against ROUGE-1, BERTScore-F1, and a MedCon-proxy (F1 over a medical word list plus dose/lab phrases, i.e., whether both notes mention the same medical concepts).

\textbf{C4:} We correlate the objectified rubric with user ratings that the judges never see. On $1{,}446$ rated production notes, mean instance-rate burden falls at every rating step. The association is small, and a named complaint does not track its mapped judge.

\section{Method}
\label{sec:method}
This section describes how we go from user feedback to a set of evaluators that judge note quality (Figure~\ref{fig:method}).

\subsection{Rubric induction and judgeability triage}
Themes are clustered from clinician comments on generated notes. We do not claim mining as novel. Two claims matter: naive frequency is not a ranking (Table~\ref{tab:c1-themes}), and the enforceability filter then deletes the frequent composites. Each theme receives one of three dispositions:

\begin{itemize}[nosep, leftmargin=*]
\item \textbf{Judgeable} from the note, transcript, and generation context alone.
\item \textbf{Needs context beyond the generation artifacts.} For example, judging generated ICD codes requires billing expertise the artifacts do not carry.
\item \textbf{Not a documentation defect}, but a software or UI issue not attributable to the generator. Excluded from the rubric.
\end{itemize}

\subsection{Objectification}
Judgeable themes are then restated in a form an LLM can decide unambiguously. We illustrate with {\em Misplacement}, one of the most common themes, which users report as content appearing in the wrong section, such as A\&P content sitting in HPI, or the reverse. It is not monolithic but composed of sub-metrics, so scoring it directly invites ambiguous interpretation. Four operations split it into determinate metrics:

\textbf{Reduction by logical decomposition:} Misplacement is essentially content appearing in a wrong section and either absent from the correct one, or present. We term content present in the wrong section as {\em Leakage}. If the content is also present in the correct section, the duplicate is redundant; if it is absent there, it is an Omission. Misplacement is therefore a composite of determinate checks. Cross-section duplication is scored through leakage, so the Redundancy judge stays within one section (scope restriction, below).

\textbf{Directional sub-categorization:} Leakage can be further split into simpler sub-categories consisting of ordered section pairs (source, destination) indicating the direction of leakage. These pairs are judged independently, for example Plan into History or Results into Physical Exam. This level of detailed enumeration is required since each of these pairs has specific rules for making the judgment. For example, for {\em A\&P in History}, if a History sentence contains a definitive diagnosis or impression, it is a violation. On the other hand, symptom narrative, patient quotes, or brief non-decisional history are acceptable.

\textbf{Scope restriction:} Some metrics also need an unambiguous scope. Redundancy means duplication within a section. Duplication across sections is unstable across reviewers and is covered by the leakage checks.

\textbf{Preference parameterization:} Even sectional complaints are sometimes preferences rather than defects (\textit{``put the plan in the HPI''} or \textit{``merge ROS into History''}). We do not treat those as leakage. 

\subsection{Evaluator battery}
With the rubric fixed, we build one judge per metric by combining a prompt with a foundation model. The prompt for each metric captures its specific nuances, which need to be handled separately. We describe their structure and principles but do not release the prompts (the released labels, ACI-Defect, criterion strings, and the planted-span hit rule support reproduction; \S\ref{sec:experiments}). Evaluator prompts follow the construction rules below. Deployed groups are in Table~\ref{tab:battery-groups}. 

\begin{itemize}[nosep, leftmargin=*]
\item \textbf{One judge per criterion, instance grain.}
Each criterion is judged by an independent agent giving structured verdicts on the constituents of a clinical note (sentences, case facts, or occurrences), rather than judging at a macroscopic level. Each verdict includes a label, supporting note and transcript excerpts, rationale, and severity (low, medium, or high clinical consequence).
\item \textbf{Anchor-then-classify.}
The prompt requires the judge to first extract intermediate artifacts (such as note or transcript quotes) before making the final judgment in accordance with the principle of providing \textit{proof before preference} \citep{tapwal2026faithful}.
\item \textbf{Dual ground truth, transcript wins.}
LLM evaluators receive the raw transcript as the primary source and the distilled atomic clinical case facts generated by an LLM from the transcripts as a secondary backup, with the transcript taking priority in case of conflict. They also receive the encounter metadata.

\end{itemize}

\section{Experimental Setup}
\label{sec:experiments}
\textbf{Two corpora:} Judges were developed on notes from a commercially deployed ambient scribe; those notes contain PHI and are not released. ACI-bench (207 pairs, \texttt{test1} $n{=}40$, CC BY 4.0) is the public test. It has no ratings or comments, so C1 aggregates and the production rating analysis cannot run on it.

\textbf{Judge model:} \texttt{gemini-2.5-pro}, constrained structured output with schema validation and repair-on-failure. As a check against prompt overfit, the ``History of'' misuse judge was re-run unchanged on a second frontier model and agreed with the original verdicts on over $98\%$ of instances ($n{=}100$). The robustness runs (Table~\ref{tab:matched}) also use \texttt{gpt-4.1} and \texttt{qwen3-next-80b}, at temperature 0 with the same hit rule.

\textbf{Cost:} One constrained-decoding call per criterion per note, one case-fact extraction, and one hallucination call per 20 sentences, so about 14 calls per note for the 11 LLM judges deployed on ACI-bench. The deterministic checks add none, and calls are independent across notes and criteria. Objectification sets that cost. The broad complaint in Table~\ref{tab:ablation} is a single call. Each section-pair check is one call. Under matched effort, the four directional calls cost 4.2$\times$ one composite call on Gemini (USD 24.71 vs 5.90 per 1{,}000 notes).

\textbf{Labels on gold (C2):} Clinicians manually reviewed all 40 notes of the ACI-bench MEDIQA-Chat Task B split (\texttt{test1}) and scored them on the grounding dimensions. Each reviewer saw the transcript, the note, the patient and EHR metadata, and the LLM-extracted case facts. Table~\ref{tab:c2} reports judges' P/R against the raw independent sheet \emph{and} against corrected gold. The labeling pass was blinded to judge output; the discrepancy adjudicator was not. Adjudication covers every discrepancy (150/150 hallucination; 280/280 omission).

\textbf{ACI-Defect (C3):} Generic perturbations (delete / repeat / shuffle) do not test whether an evaluator detects the defect it was designed for, nor whether reference metrics would have caught it. We plant one defect at a time into the 40-note ACI-bench \texttt{test1} split, leaving all other content untouched (notes are not re-rendered). If a note lacks the section a plant needs, we skip the plant rather than invent a section. A \textbf{hit} is a judge instance whose snippet overlaps the planted span and whose class is the plant's target class; it is not a note-level failure verdict. Against each perturbed note we also score ROUGE-1/2/L F1, BERTScore-F1 (roberta-large, library default), and a UMLS-free \textbf{MedCon-proxy} (gazetteer + dose/lab concept F1), all versus the \textit{original gold}, reporting drops as $1-\text{score}$. Recipes: Table~\ref{tab:plant-catalog}.

\textbf{Planted defects are cleaner than production defects.}
The same leakage judges, mapped onto clinician misplacement labels rather than onto a planted span, score 0.24 precision and 0.13 recall (Appendix~\ref{app:c2}). That mapping is not a quality claim for Table~\ref{tab:c2}. It is evidence that a surgical plant is easier than the misplacement clinicians actually mark, which is often ``plan belongs in assessment'' rather than one moved sentence.

\textbf{Production notes--ratings correlation (C4):} After dropping EHR/UI comments, $1{,}446$ notes carrying a user rating are judged. Comments, ratings, and user identifiers never enter the evaluator. We score \emph{instance rates} (target findings / opportunities) rather than note-level pass/fail (for example, inappropriately omitted case facts, fabricated sentences). A comment is a family tag, not a gold span. For each dimension we also record the note-level ``at least one finding'' bit, so rate and flag are comparable on identical inputs. \emph{Burden} is the mean of column-wise $z$-scored rates over six dimensions scored on every note; an out-of-scope dimension is skipped, not counted as zero.

\section{Evaluation}
In this section, we evaluate the different components of the evaluation pipeline described in Section~\ref{sec:method} and motivate the need for each of those components in the pipeline. We start by showing why it is necessary to objectify user feedback into determinate metrics.

\subsection{The enforceability filter discards half of the reported defects (C1)}
The complaints an LLM judge cannot grade as posed are as common as the ones it can. We took a year of production feedback: $10{,}801$ rated notes carrying $3{,}496$ free-text comments, rated 1--5 (5 = most satisfied). Table~\ref{tab:c1-themes} assigns each unfavorable comment to exactly one family. Composite complaints come from $14.3\%$ of unsatisfied users, while complaints that are already atomic or that need an external tool come from $14.4\%$. An enforceability filter deletes the first group and keeps most of the second, so it discards about half of everything users report as a note defect.

\begin{table}[t]
  \caption{Complaint families from a year of production feedback, user-weighted
    over the $480$ users who left an unfavorable comment (rating ${<}4$ and more than two words). Families are disjoint and sum to $100$\% up to rounding; the
    indented row is the largest tag \emph{within} composites, not a sixth
    family. The keyword audit is a lower bound, so the unclassified residual is
    not a claim that those comments are defect-free.}
  \label{tab:c1-themes}
  \centering
  \small
  \setlength{\tabcolsep}{5pt}
  \renewcommand{\arraystretch}{0.95}
  \setlength{\aboverulesep}{0.25ex}
  \setlength{\belowrulesep}{0.25ex}
  \begin{tabularx}{\linewidth}{@{}
      >{\raggedright\arraybackslash}p{0.27\linewidth}
      r
      >{\raggedright\arraybackslash}X@{}}
    \toprule
    \textbf{Family (disjoint)} & \textbf{\%} & \textbf{Disposition} \\
    \midrule
    Composite structure & \textbf{14.3}
      & Unenforceable as posed; objectified into leakage, format, redundancy \\
    \quad of which misplacement & 10.0
      & Four directional leakage judges \\
    Atomic / tool-needing & \textbf{14.4}
      & Judged directly, except billing codes, which need a tool \\
    Not a note defect & 6.8 & Excluded from the rubric \\
    Standing preference & 0.1 & Parameterized, not scored as leakage \\
    Unclassified & 64.5 & Keyword lower bound \\
    \bottomrule
  \end{tabularx}
\end{table}

\subsection{Quality gaps in ACI-bench (C2)}
In the previous section, we established the importance of objectifying composites to be able to preserve valuable insights gained from the feedback. In this section, we devise experiments to test the effectiveness of the rubric derived from user feedback using ACI-bench \texttt{test1} notes.

\begin{figure}[t]
  \centering
  \includegraphics[width=\linewidth]{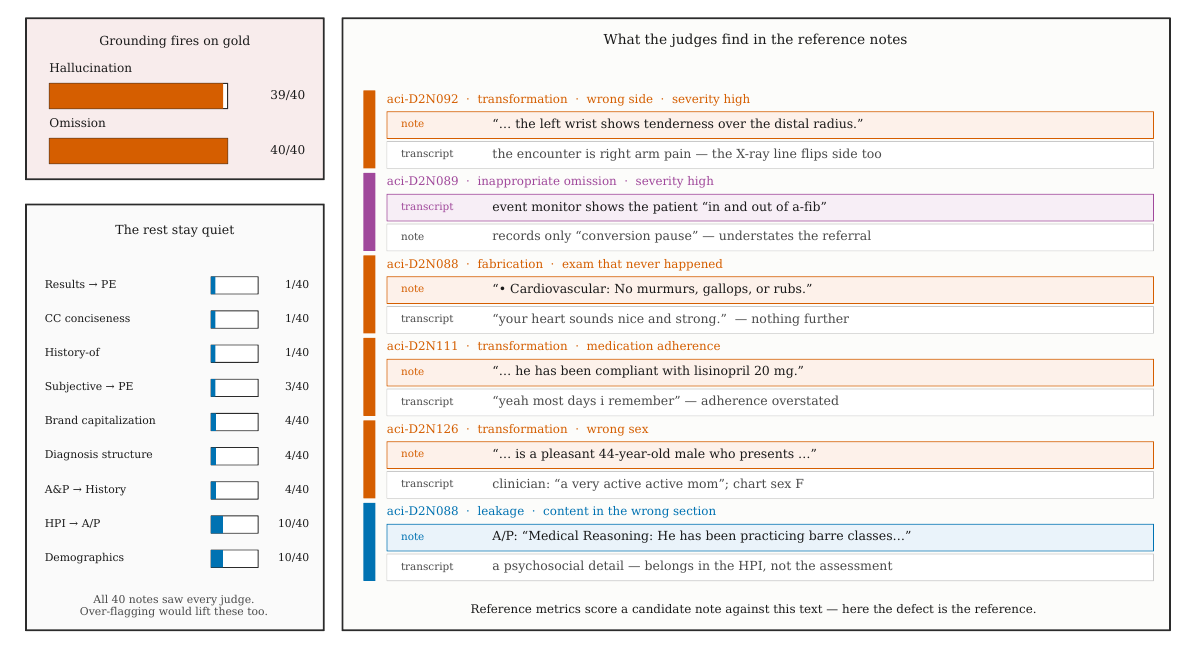}
  \caption{ACI-bench reference notes have quality gaps. Left: raw judge fire rates on unmodified gold, before adjudication. The grounding judges fire on almost every note, while leakage, language, structure, and identity stay low. (Table~\ref{tab:c2} reports the smaller \emph{confirmed} counts.) Right: six judge verdicts on \texttt{test1} gold, each pairing the note line with the transcript evidence; all grounding findings are clinician-confirmed.}
  \label{fig:gold-dirt}
\end{figure}

Before we can use ACI-bench as a gold benchmark for our experiments, we need to ensure that it does not suffer from any quality gaps as defined by the user-feedback-derived rubric. This is inspired by anecdotal evidence in ACI-bench. For example, in \texttt{aci-D2N092} the gold documents ``Examination of the \textbf{left} wrist shows tenderness over the distal radius'' for an encounter whose chief complaint is \textbf{right} arm pain. Figure~\ref{fig:gold-dirt}(right) gives more such examples. Run on all $40$ notes, the grounding judges flag hallucination and omission on almost every note, while the other judges fire rarely (Figure~\ref{fig:gold-dirt}, left). 

We do not treat these flags as confirmed errors until a clinician has ruled on them, and we do not edit ACI-bench. Note text is byte-identical to the original corpus \citep{yim2023acibench} throughout; we only change the labels associated with each note. A label is a clinician decision on one sentence or one case fact
(supported vs not; missing vs not). It is not an edit to the note text. The sheet design, the adjudication rules, and the residual false positives are in Appendix~\ref{app:c2}. Three clinicians partitioned the note set and marked every sentence and every transcript-derived case fact with judge output hidden ($1{,}878$ sentences; $2{,}062$ facts); each item carries one independent label, and every judge-labeler discrepancy was then adjudicated. Table~\ref{tab:c2} reports judge precision and recall. Adjudication more than doubles the hallucination positive class (44 $\to$ 102 sentences) and moves omission from 334 to 368 facts. The corrected labels state the condition of the benchmark precisely: $102$ of $1{,}878$ gold sentences ($5.4\%$) carry a confirmed hallucination and $368$ of $2{,}062$ case facts ($17.8\%$) are inappropriately missing. Every one of the $40$ reference notes omits at least one discussed fact, and $32$ of $40$ contain at least one hallucinated sentence.

A second clinician, who did not do the first pass and did not see judge output, relabeled $10$ of the $40$ notes ($915$ items). Note-level agreement is $10/10$ for both dimensions. Span-level Cohen's $\kappa$ is $0.61$ for omission ($95\%$ CI $0.49$--$0.71$) and $0.41$ for hallucination ($0.22$--$0.58$), against first-pass prevalence of $11.0\%$ and $4.7\%$. The second rater marked more defects than the first, so judge precision against the first pass is a floor. This study does not validate the adjudicated labels. We release corrected labels rather than a cleaned corpus, because automated repair of the reference notes would make gold an output of the system being evaluated against it.

\begin{table}[t]
  \caption{Objectified judges vs gold on $n{=}40$ \texttt{test1} originals, before and after
    discrepancy adjudication. Instance grain. Hallucination = binary
    has-hallucination (any of fabrication / transformation / inference /
    misattribution; benign-only excluded). Omission = inappropriately missing or
    partially so. \emph{Notes} = of 40 with $\ge 1$ positive. Adjudicator was not
    blinded: corrected precision is an upper bound.}
  \label{tab:c2}
  \centering
  \small
  \setlength{\tabcolsep}{5pt}
  \begin{tabular}{@{}llcccrc@{}}
    \toprule
    Dimension & Units & Raw P / R & Corr. P / R & F1 raw $\to$ corr.
      & Positives & Notes \\
    \midrule
    Hallucination & sentences & 0.21 / 0.86 & \textbf{0.57 / 0.97}
      & 0.34 $\to$ 0.72 & 44 $\to$ 102 & 32 \\
    Omission & case facts & 0.56 / 0.79 & \textbf{0.66 / 0.83}
      & 0.66 $\to$ 0.74 & 334 $\to$ 368 & 40 \\
    \bottomrule
  \end{tabular}
\end{table}

\subsection{Judges find the planted defect where reference metrics cannot (C3)}
To test whether a judge finds the violation it was built for, we plant one defect at a time into gold \texttt{test1} and call the resulting dataset \emph{ACI-Defect}; Figure~\ref{fig:p1-schematic} (Appendix~\ref{app:battery}) shows one such plant. Flags that miss the plant are the false-positive baseline already measured in Figure~\ref{fig:gold-dirt}(left).

\begin{table}[t]
  \caption{Planted-span recall vs reference-metric drop. Recall = a judge instance overlapping the planted span with the plant's target class; not note-level \texttt{is\ub failure}, not F1. The last three columns are drops ($1-\text{score}$) vs original gold, self-similarity ceiling $1.0$. $n/40$ = notes planted. Every plant is localized while reference metrics barely move. P2 and P8 are lowest because of a known extractor window and a gazetteer coverage gap in those two checks, not because the plant is subtler.}
  \label{tab:plants}
  \centering
  \footnotesize
  \setlength{\tabcolsep}{3pt}
  \begin{tabularx}{\linewidth}{@{}l L c L r r r r@{}}
  \toprule
    ID & Plant & $n/40$ & Target class & Recall & ROUGE-1 & BERTScore & MedCon-p \\
    \midrule
    P1  & A\&P sentence $\to$ History  & 28/40 & \texttt{leakage}
        & 0.96 (27/28) & 0.000 & 0.017 & 0.006 \\
    P2  & HPI narrative $\to$ A\&P     & 28/40 & \texttt{leakage}
        & 0.54 (15/28) & 0.000 & 0.017 & 0.000 \\
    P3  & Symptom sentence $\to$ PE   & 40/40 & \texttt{leakage}
        & 1.00 (40/40) & 0.023 & 0.010 & 0.000 \\
    P4  & RESULTS sentence $\to$ PE   & 30/40 & \texttt{leakage}
        & 0.90 (27/30) & 0.018 & 0.009 & 0.000 \\
    P5  & Fabricated normal PE finding & 40/40 & \texttt{fabrication}
        & 1.00 (40/40) & 0.016 & 0.011 & 0.000 \\
    P6  & Delete modifier span        & 36/40 & \texttt{omission}
        & 0.83 (30/36) & 0.007 & 0.005 & 0.037 \\
    P7  & \textit{History of} on active problem & 33/40 & \texttt{History of misuse}
        & 0.94 (31/33) & 0.003 & 0.002 & 0.000 \\
    P8  & Lowercase brand names       & 19/40 & \texttt{uncapitalized\ub drug}
        & 0.68 (13/19) & 0.000 & 0.001 & 0.000 \\
    P9  & Bundle two A\&P headings     & 19/40 & \texttt{bundled diagnosis}
        & 0.95 (18/19) & 0.002 & 0.004 & 0.000 \\
    P10 & Rewrite CC as narrative     & 26/40 & \texttt{too many words}
        & 1.00 (26/26) & 0.030 & 0.023 & 0.000 \\
    P11 & Swap sex vs chart           & 30/40 & \texttt{sex / chart mismatch}
        & 1.00 (30/30) & 0.004 & 0.001 & 0.000 \\
    \bottomrule
  \end{tabularx}
\end{table}

\begin{table}[t]
\centering\small
\setlength{\tabcolsep}{4pt}
\caption{Composite versus objectified checks on four misplacement plants.
All rows share the model, inputs, decoding, schema, and hit rule; the objectified row is the deployed multi-step judge, so prompt effort also differs (Table~\ref{tab:matched}).
\emph{Complaint, as posed} is ``Content is in the wrong section of the note.''
\emph{Clean} is the share of the 40 unperturbed notes flagged; $J$ uses P2--P4.}
\label{tab:ablation}
\begin{tabular}{@{}lcccc@{\hspace{9pt}}ccc@{}}
\toprule
& \multicolumn{4}{c@{\hspace{9pt}}}{Planted-span hits / $n$}
& \multicolumn{3}{c}{As an instrument} \\
\cmidrule(r){2-5}\cmidrule(l){6-8}
Criterion & P1 & P2 & P3 & P4 & Recall & Clean & $J$ \\
\midrule
Objectified pairs      & 27/28 & 15/28 & 40/40 & 27/30 & 0.87 & 0.02--0.26 & \textbf{0.72} \\
Complaint, as posed    & 28/28 & 28/28 & 40/40 & 30/30 & \textbf{1.00} & 0.70 & 0.30 \\
Zhou et al.\ criterion & 26/28 & 27/28 & ---   & ---   & 0.95 & --- & --- \\
\bottomrule
\end{tabular}
\end{table}

Table~\ref{tab:plants} is the result. Every plant is localized by its target judge at a median recall of $0.95$, while ROUGE-1, BERTScore-F1, and the MedCon-proxy move by at most $0.037$; ROUGE-1 moves by at most $0.004$ on the four plants that leave the token bag unchanged or swap a single token (P1, P2, P8, P11). For scale, deleting $15\%$ of tokens at random drops ROUGE-1 by $0.081$. The two lowest recalls are limits of those checks rather than subtler plants: P2's extractor does not read past an \texttt{INSTRUCTIONS} header, costing 11 spans, and P8's gazetteer lacks six brand names.

\textbf{Objectification improves discrimination, not recall:} A natural objection is that objectification is unnecessary, and we can simply give a frontier model the complaint and it will find the defect. It does. Given only ``Content is in the wrong section of the note.'', the same judge localizes the planted span in all 126
planted notes, beating the 109/126 of the directional-pair judges, and the criterion \citet{zhou2025feedback} discard scores 53/56 on the pair it covers (Table~\ref{tab:ablation}). That judge is nonetheless useless as an instrument, because it also fires on clean notes. It flags 28 of the 40 unperturbed notes ($70\%$), at $1.60$ flagged sentences per note against $1.68$ on planted notes, so its output cannot tell a note with a surgical defect from a clean one. The directional judges flag the same notes at $2$--$26\%$. Scoring each version by recall minus false-positive rate (Youden's $J$, where $1.0$ is perfect and $0$ is uninformative), objectification reaches $0.72$ against the composite's $0.30$, though this comparison also differs in prompt effort, which Table~\ref{tab:matched} controls. Part of the composite's failure is scope, not capability. Half of the composite's clean-note flags are content the objectified rubric explicitly allows in History: pertinent negatives that belong to Review of Systems, and past, family and social history (Appendix~\ref{app:ablation}). Asked an unscoped question, the judge enforces an idealized template of its own instead of the ordered section pair the complaint names.

\begin{table}[t]
\centering\small
\setlength{\tabcolsep}{4pt}
\caption{Matched effort on P2--P4. Both versions use the same anchor-then-classify procedure, the same allowed and disallowed rules, and the same three calibration examples; only the question differs. Mean FPR averages per-check clean-note rates. Union FPR fails a clean note if any objectified check fires.}
\label{tab:matched}
\begin{tabular}{@{}lcccc@{}}
\toprule
Model & Obj.\ $J$ (mean FPR) & Composite $J$ (FPR) & Obj.\ $J$ (union FPR) & P2 $J$ obj / cmp \\
\midrule
Gemini 2.5 Pro  & \textbf{0.81} (0.18) & 0.62 (0.38) & 0.51 (0.47) & 0.60 / 0.62 \\
GPT-4.1         & \textbf{0.66} (0.22) & 0.40 (0.53) & 0.34 (0.55) & 0.19 / 0.22 \\
Qwen3-Next-80B  & \textbf{0.67} (0.23) & 0.45 (0.51) & 0.28 (0.62) & 0.33 / 0.49 \\
\bottomrule
\end{tabular}
\end{table}

\textbf{Under matched effort the gain is per check, not per note.} When both versions use the same procedure, rules, and examples, one directional question per call still lowers the per-check clean-note false-positive rate on all three models, and $J$ rises by $0.19$, $0.26$, and $0.22$ (Table~\ref{tab:matched}). The gain comes from the two Physical Exam checks; HPI$\to$A\&P shows no advantage on any model. When the checks are combined into one note-level decision, the single composite question is the better screen on all three models. Objectification buys a localizable per-criterion judgment, not a better note-level alarm. Much of the original Gemini gap ($0.72$ vs.\ $0.30$) was prompt effort. Called once, without the deployed extractor, the objectified P2 predicate recalls $28/28$, so the $15/28$ in Table~\ref{tab:ablation} is an extractor limit.

\textbf{The decomposition matters.} Rewriting the same complaints by content class (a management decision outside A\&P, symptom narrative outside History, a result outside Results) instead of by ordered section pair keeps recall (pooled P2--P4: $0.91$, $0.87$, $0.91$ on Gemini, GPT-4.1, Qwen) but not false-positive control (mean clean rate $0.67$, $0.54$, $0.39$; $J$ $0.24$, $0.33$, $0.52$). Finding the plant is robust to the decomposition; the low false-positive rate is specific to the directional split.

\textbf{Blind spots}, found through plant misses, judge--human adjudication, and cross-model reruns: beyond the P2 extractor window, the P8 gazetteer gap, and the composite's self-imposed template, cross-model runs exposed two more: the scoped fabricated-exam predicate recalls $0.00$ on GPT-4.1 and Qwen, and the objectified ``history of'' check flags $57\%$ of clean notes on Qwen.

\begin{figure}[t]
  \centering
  \includegraphics[width=0.95\linewidth]{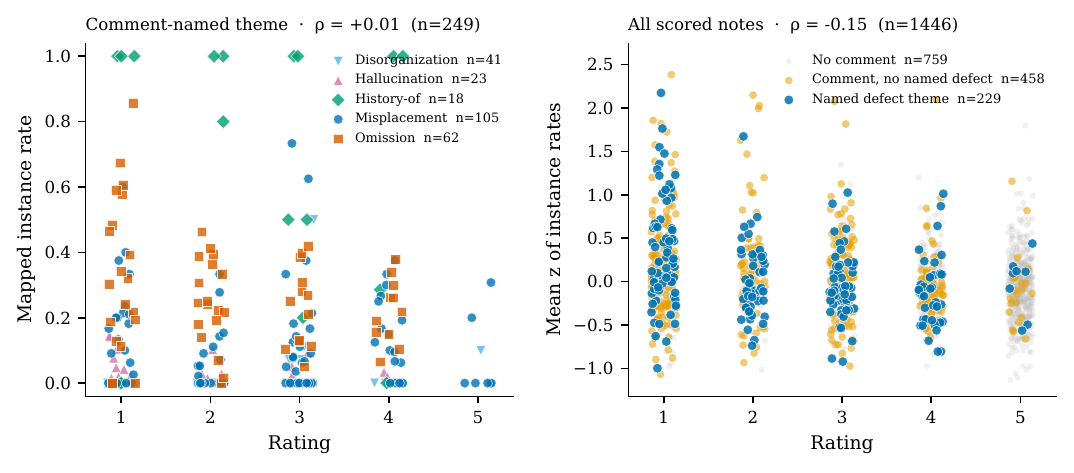}
  \caption{Objectified rates agree with the rating in aggregate, not on the
  named family. Left: only notes whose comment names a defect
  ($n{=}249$ pairs / 229 notes); notes with no comment and comments that
  name no defect (praise such as ``good'', or otherwise unclassified)
  are absent by construction. $y$~=~instance rate of the mapped judge.
  $\rho{=}{+}0.01$. Right: all scored notes ($n{=}1{,}446$). At rating 5, $484/516$ notes have no comment and $8$ name a defect.
  $y$~=~mean $z$ of instance rates. $\rho{=}{-}0.15$ overall;
  $\rho{=}{-}0.30$ on named-defect notes. No comment text is shown.}
  \label{fig:comment-rating}
\end{figure}

\subsection{Objectified rates agree with user ratings in direction (C4)} 
C1--C3 never use a human judgment of a whole note. For that we use the optional 1--5 rating clinicians leave in production. Ratings, comments, and user identifiers are hidden from every judge, and the comment-theme-to-judge map was fixed before scoring (Appendix~\ref{app:comment-theme}).

\textbf{Rates rank notes; flags do not.} A note-level ``at least one judge fired'' bit is useless here: it fires on $86\%$ of notes for hallucination and on $97\%$ for both omission and diagnosis structure, so nearly every note looks alike. The same judges scored as rates do separate the rating bands. Averaging each note's standardized rates over the six dimensions scored on every note
gives one \emph{burden} score, and mean burden falls from $+0.15$ at rating 1 to $-0.08$ at rating 5, dropping at every step (Figure~\ref{fig:comment-rating}, right; $\rho{=}{-}0.15$, $p{<}10^{-7}$, $n{=}1{,}446$; $\rho{=}{-}0.30$ for overall burden on the $229$ named-defect notes). Five of the six dimensions are lower at rating 5 than at rating 1. The effect is small. Since no judge sees the rating, this is convergent evidence of direction only, not a basis for routing notes or validation of any one judge.

\textbf{A note-level score cannot name the criterion.} One number per note cannot say which of $k$ criteria was violated. The data bear this out: when a comment names a family, the matching instance rate is not higher, except for bundled diagnosis headings; omission, A\&P$\to$HPI leakage, ``history of'' misuse, and hallucination show no enrichment (Table~\ref{tab:c4-rates}), and the rating does not correlate with the mapped judge at all ($\rho{=}{+}0.01$, $n{=}249$; Figure~\ref{fig:comment-rating}, left). Rating-5 notes are not defect-free either: they still omit about $21\%$ of case facts and leave about $6\%$ of sentences unsupported. This is why we test localization on planted spans:
production feedback names a family, never a span.

\section{Conclusion}
When evaluation criteria are mined from user feedback, the enforceability filter is biased against the most common complaints. Objectification is the alternative to deletion. Posed as a composite, the same judge localizes every planted misplacement yet fires on 70\% of clean reference notes. With prompt effort matched, objectified checks still lower per-check false positives on three models, but a single composite remains the better note-level screen, so the gain is localization per criterion. Across eleven plants, objectified judges localize the planted span (median recall $0.95$, range $0.54$--$1.00$) while ROUGE-1 drops by at most $0.030$. Since production notes cannot be published under HIPAA, ACI-bench is the public test. Because it contains quality gaps under the same rubric, we release independent clinical review labels and grounding judges' performance on it. On $1{,}446$ rated production notes, the same objectified rates move in the right direction against a clinician rating no judge sees, with the mean falling at every step from rating 1 to rating 5. The association is small and is not a per-criterion validation. That check costs no annotation but works only at the right grain: a note-level pass/fail bit fires on most notes and ranks nothing. Score defects per opportunity, use user feedback to decide \emph{what} to measure, validate localization on planted spans, and do not treat ACI-bench gold as a clean negative set.

\section{Limitations}
Production data are from one site, and ratings are optional and self-selected, so $\rho{=}{-}0.15$ describes this feedback channel, not clinician-perceived quality in general. We fit no classifier and do not use burden to route notes. The ACI-bench review covers 40 \texttt{test1} notes. Agreement was measured on 10 doubly labeled notes; span-level $\kappa$ for hallucination is moderate. Adjudication was not blinded, so corrected precision in Table~\ref{tab:c2} is an upper bound, and it certifies grounding only. Objectification is an expert restatement by one clinician. We did not measure agreement between independent rewriters; the content-class rewrite is a proxy, and it keeps recall but not false-positive control. Matched effort covers three models and P2--P4 only. We did not test the order or presentation of criteria within a prompt \citep{shankar2024validators}. Clean-note flags are all scored as false positives, though some are real reference-note defects and some are documentation convention. Plant-level recall does not transfer to production misplacement labels (0.24 precision, 0.13 recall; Appendix~\ref{app:c2}). The 11-plant battery uses Gemini 2.5 Pro with the deployed prompts, which are not released; the composite, matched-effort, and content-class criterion strings are released.

\section{Ethics}
The production study uses de-identified encounter aggregates under HIPAA policy. Notes, transcripts, and verbatim complaints are PHI and appear in neither the paper nor the release. ACI-bench is an existing public corpus, and ACI-Defect and the independent labels add no new patient data.

\bibliographystyle{plainnat}
\bibliography{references}

\clearpage
\appendix


\section{Instruments and evaluator battery}
\label{app:instruments}

\noindent\textit{Supports Section~1 and Section~3.3.} 

Table~\ref{tab:instruments} contrasts what each class of instrument can detect for the defects clinicians raise; Table~\ref{tab:battery-groups} lists the deployed judge groups.

\begin{table}[H]
  \caption{What each instrument can see (complaint-derived defects; not a bake-off).}
  \label{tab:instruments}
  \centering
  \small
  \setlength{\tabcolsep}{4pt}
  \renewcommand{\arraystretch}{0.88}
  \setlength{\aboverulesep}{0.25ex}
  \setlength{\belowrulesep}{0.25ex}
  \begin{tabularx}{\textwidth}{@{}
      >{\centering\arraybackslash}p{0.18\textwidth}
      >{\centering\arraybackslash}X
      >{\centering\arraybackslash}p{0.17\textwidth}
      >{\centering\arraybackslash}p{0.18\textwidth}
      >{\centering\arraybackslash}p{0.15\textwidth}@{}}
    \toprule
    \textbf{Defect clinicians raise}
      & \textbf{Example (ACI-bench / synthesized)}
      & \textbf{PDQI-9 / PDSQI-9}
      & \textbf{ROUGE / MedCon vs gold}
      & \textbf{Objectified check} \\
    \midrule
    Assessment sentence in History.
      & ``I recommend \ldots\ Robitussin'' in History
      & No (not sectional leakage)
      & No if tokens unchanged
      & Directional leakage \\
    \midrule
    Fabricated normal exam.
      & ``CN II--XII intact'' inserted into PE
      & Partial (accuracy)
      & Weak if few tokens
      & Sentence hallucination \\
    \midrule
    Discussed dose missing from plan.
      & Spoken dose change omitted
      & Partial (completeness)
      & Weak if other overlap high
      & Case-fact omission \\
    \midrule
    ``History of'' on an active problem.
      & Active diagnoses rewritten ``History of X''
      & No
      & Near-paraphrase
      & ``History of'' misuse \\
    \midrule
    Documented sex $\neq$ chart.
      & ``59-year-old male'' $\to$ female; chart stays M
      & No
      & One-token swap
      & Demographics \\
    \bottomrule
  \end{tabularx}
\end{table}

\begin{table}[H]
  \caption{Evaluator battery. Each group is an independently deployable
    sub-agent. ACI-bench has no preference records and no billing context,
    so style judges run under documented defaults and the diagnostic-code
    judge is omitted.}
  \label{tab:battery-groups}
  \centering
  \small
  \setlength{\tabcolsep}{4pt}
  \begin{tabularx}{\linewidth}{@{}
      >{\centering\arraybackslash}p{0.18\linewidth}
      >{\centering\arraybackslash}X
      >{\centering\arraybackslash}p{0.28\linewidth}@{}}
    \toprule
    \textbf{Group} & \textbf{Evaluators} & \textbf{On ACI-bench?} \\
    \midrule
    Grounding
      & Omission (case-fact level), hallucination (sentence level)
      & Yes \\
    Directional leakage
      & A\&P \arr{} History, HPI \arr{} A\&P,
        subjective \arr{} PE, results \arr{} PE
      & Yes \\
    Language use
      & ``History of'' misuse, provider voice
      & Yes (defaults) \\
    Structure
      & Diagnosis structure, within-section redundancy
      & Yes \\
    Identity
      & Demographics documentation
      & Yes (age / sex; no stored gender-identity prefs) \\
    Deterministic
      & Regex/rule style, verbosity bands, section presence, CC conciseness, section format
      & Yes \\
    Tool-equipped
      & Diagnostic code existence / mapping
      & No --- no billing context \\
    \bottomrule
  \end{tabularx}
\end{table}


\section{Clinician labels on ACI-bench gold}
\label{app:c2}

\noindent\textit{Supports Section~5.2 and Table~\ref{tab:c2}.}

\paragraph{Sheet design.}
Each independent sheet has one row per note sentence, labeled with one of four hallucination types or benign, plus composite misplacement, redundancy, and flow, and one row per generated case fact, labeled for missingness. History-of misuse and demographics were not labeled. Sheet rows are matched to judge output by line number or exact text. Sheet comments are incomplete and are not used for precision or recall.

\paragraph{Adjudication.}
Every judge--sheet discrepancy was ruled on (150 hallucination, 280 omission) under the rules in Table~\ref{tab:adj-rules}. Hallucination is scored as a binary has-hallucination decision.

\begin{table}[H]
  \caption{Discrepancy adjudication rules.}
  \label{tab:adj-rules}
  \centering
  \small
  \begin{tabular}{@{}ll@{}}
    \toprule
    Verdict & Corrected gold label \\
    \midrule
    Judge correct & Follows the judge \\
    Independent label correct & Follows the sheet \\
    Both wrong & Negative \\
    Grain / uncertain & Dropped (11 hallucination sentences; 8 omission facts) \\
    \bottomrule
  \end{tabular}
\end{table}

\paragraph{Residual false positives.}
Eight rows confirm a hallucination while disputing its type; under binary scoring they are true positives for the judge (precision $0.52 \to 0.57$; recall unchanged). After that credit, 75 residual hallucination false positives remain. They are mostly licensed clinical summarization (the normal-exam convention, plural-to-bilateral rewording, and HPI paraphrase of the patient's account); 6 are ASR or diarization errors charged to the model (e.g., a 4\,mg Lasix dose that does not exist), and 4 are flow or omission defects flagged under the hallucination head.

\paragraph{Misplacement overlay.}
Mapping the four leakage judges onto clinician misplacement labels gives 0.24 precision and 0.13 recall (46 clinician-positive sentences on 16 notes, after dropping intra-subjective ROS/HPI moves; 32 of the 46 are ``plan belongs in assessment''). That mapping tests the wrong object and is \textbf{not} a Table~\ref{tab:c2} quality claim.


\section{ACI-Defect plant catalog}
\label{app:battery}

\begin{figure}[H]
  \centering
  \includegraphics[width=\linewidth]{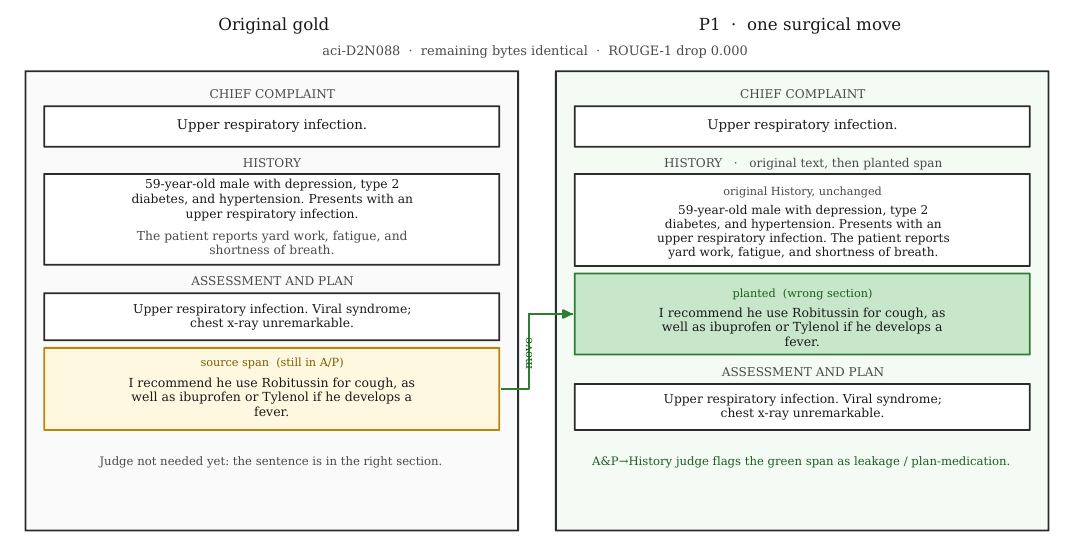}
  \caption{An ACI-Defect plant (P1): one A\&P sentence moved verbatim into History; all other bytes unchanged, ROUGE-1 drop $0.000$.}
  \label{fig:p1-schematic}
\end{figure}

\noindent\textit{Supports Section~5.3 and Table~\ref{tab:plants}.} Table~\ref{tab:plant-catalog} gives each plant's recipe and target judge; Figure~\ref{fig:p1-schematic} shows P1 on one note.

\begin{table}[H]
  \caption{The ACI-Defect battery. Each perturbation is a determinate,
    localizable edit with a single target evaluator. The right column states
    why reference metrics are expected to be insensitive: most plants leave
    the token bag or the concept set unchanged, so ROUGE and MedCon-proxy
    have little to detect even when the clinical meaning changes. P12 ships
    with the overlay but carries no recall in Table~\ref{tab:plants}: its
    first-person cue already occurs in the reference notes, so a flag cannot
    be attributed to the plant.}
  \label{tab:plant-catalog}
  \centering
  \small
  \setlength{\tabcolsep}{4pt}
  \begin{tabularx}{\linewidth}{@{}l A B C@{}}
    \toprule
    ID & Perturbation & Target evaluator & Why ROUGE / MedCon-proxy should be weak \\
    \midrule
    P1  & Move an A\&P sentence into History verbatim
        & A\&P \arr{} History leakage & Token bag unchanged \\
    P2  & Move an HPI narrative sentence into A\&P
        & HPI \arr{} A\&P leakage & Token bag unchanged \\
    P3  & Insert a symptom-report sentence into PE
        & Subjective \arr{} PE leakage
        & $\sim$17 tokens into a 300--700 word note \\
    P4  & Copy a RESULTS sentence into PE
        & Results \arr{} PE leakage
        & Tokens duplicated; concept set unchanged \\
    P5  & Insert a fabricated normal exam finding (e.g., cranial nerves
          II--XII intact)
        & Hallucination
        & Adds plausible clinical tokens; concept F1 may \emph{increase} \\
    P6  & Delete every copy of a case-fact modifier span (dose,
          laterality\,+\,anatomy)
        & Omission & Remaining lexical overlap high \\
    P7  & Rewrite an active A\&P problem heading as ``History of X''
        & ``History of'' misuse & Near-paraphrase \\
    P8  & Lowercase brand drug names
        & Provider voice\newline{\footnotesize\texttt{uncapitalized\_drug}}
        & Case-only; invisible to concept F1 \\
    P9  & Merge two A\&P problem headings with ``and''
        & Diagnosis structure & Concept set unchanged \\
    P10 & Rewrite CC as a full narrative sentence
        & CC conciseness & Adds tokens \\
    P11 & Swap documented sex in \texttt{N-year-old} lines; chart metadata
          unchanged
        & Demographics & One-token swap \\
    P12 & Insert a distinctive clinician-\texttt{I} sentence after the first
          header
        & Provider voice\newline{\footnotesize\texttt{first\_person\_i}}
        & One short sentence; gold already uses \texttt{I} \\
    \bottomrule
  \end{tabularx}
\end{table}

\paragraph{Random-deletion control.}
CTRL (not in Table~\ref{tab:plants}) deletes 15\% of whitespace tokens (seed 20260821) and has no target judge. It drops ROUGE-1 by $0.081$ and MedCon-proxy by $0.125$. ROUGE therefore does respond to random deletion, so its near-zero drops on the plants show blindness to these specific edits, not a metric that cannot move.


\section{Composite-vs-objectified ablation protocol}
\label{app:ablation}

\noindent\textit{Supports Tables~\ref{tab:ablation} and~\ref{tab:matched}.}

The composite prompt and the objectified judges see the same note and transcript, the same judge model at temperature 0, and the same instance-level response schema: each instance carries an exactly quoted note sentence, the section it sits under, a class, a rationale, and a confidence. The baseline is therefore not handicapped by a note-level yes/no output. The criterion text is the intended difference, but the objectified prompt is the deployed multi-step judge, so prompt effort also differs; the matched-effort runs (Table~\ref{tab:matched}) remove that difference. The composite criterion is stated as posed, with no ordered section pair, no scope restriction, and no inclusion/exclusion list.

\paragraph{Clean-note flags.}
Buckets are assigned automatically from the composite judge's own rationale text and are not clinician labels: assessment/plan placement 21 flags, past/family/social history 11, Review-of-Systems and pertinent negatives 11, chronic-condition narrative 10, exam placement 8, other 3. Of the 64 clean-note flags, 32 fall in the three buckets the objectified rubric rules acceptable in History, and for 11 of the 28 flagged notes every flag is of that kind.

\paragraph{Shared hit rule.}
Recall for every version is scored by one rule: a hit requires a flagged instance whose quoted span overlaps the planted span, where overlap is containment in either direction or content-token F1 $\ge 0.5$. That rule reproduces the objectified numbers in Table~\ref{tab:plants} exactly, so the two versions are comparable.


\section{Production rating analysis}
\label{app:comment-theme}

\noindent\textit{Supports Section~5.4 and Figure~\ref{fig:comment-rating}.}

The theme-to-judge map (Table~\ref{tab:theme-map}) was fixed before scoring. A rate is target findings over opportunities, not a note-level pass/fail. Comments never enter the evaluator, and EHR/UI comments are excluded.

\begin{table}[H]
  \caption{Comment theme to judge map, fixed before scoring.}
  \label{tab:theme-map}
  \centering
  \small
  \begin{tabular}{@{}ll@{}}
    \toprule
    Comment theme & Mapped judge rate (findings / opportunities) \\
    \midrule
    Omission & Inappropriately missing facts / case facts \\
    Hallucination & Unsupported sentences / sentences \\
    ``History of'' & Misuse instances / ``history of'' phrases \\
    Misplacement & A\&P$\to$HPI leakage rate \\
    Disorganization & Bundled diagnosis headings / diagnosis headings \\
    \bottomrule
  \end{tabular}
\end{table}

\paragraph{Burden.}
\emph{Burden} averages column-wise $z$-scored rates over the six dimensions scored on every note: omission, hallucination, ``history of'' misuse, A\&P$\to$HPI leakage, diagnosis structure, and subjective/results$\to$PE leakage. Bundled diagnosis headings are mapped from a comment theme but are not part of that average. Judges abstain when the phrase or section is absent, so those rates are over in-scope notes only.

\paragraph{Named comment vs matching rate.}
Table~\ref{tab:c4-rates} compares, for each named family, the matching judge's rate on notes whose comment names that family against all other notes.

\begin{table}[H]
  \centering
  \small
  \caption{Named comment vs matching instance rate on notes that have a
    usable comment and that judge. Not gold-span recall. Bundled
    headings are the only enrichment.}
  \label{tab:c4-rates}
  \begin{tabular}{@{}lrrrr@{}}
    \toprule
    Named family & $n$ & Named rate & Other rate & $\rho$ \\
    \midrule
    Omission & 62 & 0.27 & 0.25 & $+0.02$ \\
    History-of & 18 & 0.63 & 0.78 & $-0.08$ \\
    Hallucination & 23 & 0.045 & 0.085 & $-0.06$ \\
    Misplacement (A\&P$\to$HPI) & 105 & 0.109 & 0.102 & $+0.03$ \\
    Disorg.\ (bundled heading) & 41 & 0.024 & 0.015 & $+0.12$ \\
    \bottomrule
  \end{tabular}
\end{table}

\end{document}